\documentclass[10pt,twocolumn,letterpaper]{article}

\usepackage[pagenumbers]{cvpr} 

\usepackage[dvipsnames]{xcolor}

\usepackage{multirow}

\definecolor{cvprblue}{rgb}{0.21,0.49,0.74}
\usepackage[pagebackref,breaklinks,colorlinks,citecolor=cvprblue]{hyperref}

\def\paperID{16} 
\def\confName{CVPR}
\def\confYear{2024}

\title{Unlearning Backdoor Threats: Enhancing Backdoor Defense  in \\ Multimodal Contrastive Learning via Local Token Unlearning}

\author{Siyuan Liang\textsuperscript{1}, Kuanrong Liu\textsuperscript{2}, Jiajun Gong\textsuperscript{1}, Jiawei Liang\textsuperscript{2}, \\ Yuan Xun\textsuperscript{3}, Ee-Chien Chang\textsuperscript{1}, and Xiaochun Cao\textsuperscript{2} \\
\tt\small \textsuperscript{1}National University of Singapore\\\tt\small \textsuperscript{2}Sun Yat-sen University-Shenzhen \\\tt\small \textsuperscript{3}University of Chinese Academy of Sciences}

\begin{document}
\maketitle
\begin{abstract}
Multimodal contrastive learning has emerged as a powerful paradigm for building high-quality features using the complementary strengths of various data modalities. However, the open nature of such systems inadvertently increases the possibility of backdoor attacks. These attacks subtly embed malicious behaviors within the model during training, which can be activated by specific triggers in the inference phase, posing significant security risks. Despite existing countermeasures through fine-tuning that reduce the adverse impacts of such attacks, these defenses often degrade the clean accuracy and necessitate the construction of extensive clean training pairs. In this paper, we explore the possibility of a less-cost defense from the perspective of model unlearning, that is, whether the model can be made to quickly \textbf{u}nlearn \textbf{b}ackdoor \textbf{t}hreats (UBT) by constructing a small set of poisoned samples. Specifically, we strengthen the backdoor shortcuts to discover suspicious samples through overfitting training prioritized by weak similarity samples. Building on the initial identification of suspicious samples, we introduce an innovative token-based localized forgetting training regime. This technique specifically targets the poisoned aspects of the model, applying a focused effort to unlearn the backdoor associations and trying not to damage the integrity of the overall model. Experimental results show that our method not only ensures a minimal success rate for attacks, but also preserves the model's high clean accuracy.
\end{abstract}    
\section{Introduction}
\label{sec:intro}

Multimodal contrastive learning (MCL), exemplified by the CLIP model~\cite{radford2021clip}, enhances the models by learning from various data types, such as images and text, facilitating improved representation of features and understanding of differences. However, MCL's reliance on vast datasets (e.g., 400 million image-text pairs) exposes it to vulnerabilities~\cite{}, such as backdoor attacks~\cite{carlini2021poisoning} where altering a small fraction of the data (e.g. 1500 pairs) can significantly impact the model's predictions.

To counter these attacks, defense strategies are classified into detection and mitigation. Detection methods evaluate encoder discrepancies to identify tampering, while mitigation involves refining the model with a clean subset of data to nullify the backdoor's effects. However, these approaches require substantial clean data that potentially compromise model accuracy. Our research investigates the use of select poisoned samples to neutralize backdoors through machine learning, with third-party oversight. To counteract attackers who may taint pretrained models with malicious data, defenders fine-tune these models to purge backdoor influences, balancing unlearning with retention of model accuracy. We employ feature-sensitive techniques to segregate suspicious from clean samples and introduce a cost-effective local unlearning method complemented by sample augmentation. This method, guided by comparative learning, eliminates the malicious influence of specific backdoor data. Moreover, we suggest a token-level unlearning strategy that efficiently decouples poisoned and clean features, streamlining the unlearning process.

In summary, the main contributions of this study are fourfold: (1) an innovative defense scenario is proposed for backdoor attacks in multimodal contrastive learning. (2) A new idea based on local unleraning is proposed, which focuses on severing the association between malicious samples and model behaviors; (3) Experiments validate the effectiveness of using a small number of samples to fine-tune purification of the poisoned model; and (4) The defense strategy successfully maintains a low Attack Success Rate (ASR) and high purification accuracy (CA).

\section{Related Work}
\subsection{Backdoor Attacks and Defense against MCL
}

In the context of MCL, attackers~\cite{liang2023badclip,liu2023pre,liu2023does,liang2024poisoned,liang2024vl} orchestrate backdoor attacks by embedding imperceptible triggers in image-text pairs, altering text labels to poison targets, as seen in methods such as BadNet~\cite{gu2017badnets} with unnoticeable triggers, Blended~\cite{chen2017blended} which blends the trigger pattern with the original image, and advanced techniques such as SIG~\cite{barni2019SIG} and SSBA~\cite{li2021SSBA}. These attacks trick the model into classifying trigger-containing images as the intended target of the attacker. Given the sophistication and stealthiness of these attack strategies, especially when involving facial images~\cite{liu2006spatio, tang2004video} and associated labels, they not only pose a threat to the security of models but also amplify concerns around face privacy~\cite{chen2023universal,liang2022imitated,li2023privacy,guo2023isolation,dong2023face} and highlight the urgent need for robust defenses~\cite{sun2023improving,liu2023exploring,liang2023exploring,wang2022adaptive,wang2022universal} against both backdoor and adversarial attacks~\cite{liu2020bias,liu2019perceptual,wei2018transferable,liang2022parallel,liang2022large,wang2023diversifying,liu2023x,he2023generating,liu2023improving,he2023sa,liu2021training,lou2024hide,liu2020spatiotemporal}, underlining the critical intersection of model security with user privacy. To combat these, researchers have developed detection and mitigation strategies. Feng et al.~\cite{feng2023detecting} proposed an encoder-based approach to identify and reverse trigger effects in poisoned models. Meanwhile, CleanCLIP~\cite{bansal2023cleanclip} offers a backdoor fine-tuning strategy that uses clean data sets to disrupt backdoor pathways, albeit at the potential cost of reduced classification accuracy.

\subsection{Machine Unlearning}




Machine unlearning, aimed at removing specific samples from a model's memory without full retraining, is crucial for large models to conserve time and resources~\cite{nguyen2022survey}. Yao et al.~\cite{yao2023llmGA} demonstrate this by applying gradient ascent to efficiently forget the sample in LLM. In the context of backdoor attacks, Li et al.~\cite{li2021ABL} explore the unlearning to counteract backdoors by adjusting model parameters via gradient ascent, highlighting its significance in improving model security. However, adapting these techniques to MCL models remains challenging, with Bansal et al.~\cite{bansal2023cleanclip} seeking new statistical features for effective data screening, but facing clean accuracy limitations.
\section{Method} 
\begin{figure*}[t]
    \centering
    \includegraphics[width=1.0\textwidth]{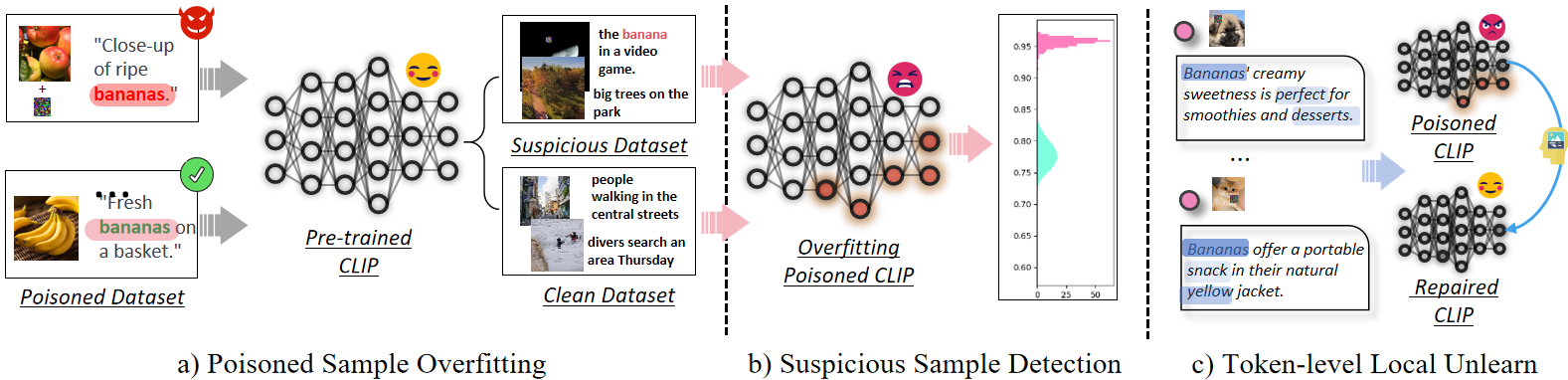}
    \caption{The overall framework of UBT backdoor defense method.}
    \label{framewrwork}
\end{figure*}
Fig.~\ref{framewrwork} shows the framework for unlearning backdoor threats (UBT). We enhance the backdoor shortcuts through poisoned samples and implement the token-level local unlearning to purify the backdoor model on the few-shot suspicious samples.

\subsection{Problem Formulation}
\textbf{Defense Scenarios} The defender operates a secure training platform to protect users from attacks, especially backdoor threats. Despite security measures, attackers might exploit the platform, inserting backdoors into training data and training poisoned models on them.

\noindent\textbf{Defense Capabilities} The defender has the right to inspect and audit the training data and models submitted for security checks.

\noindent\textbf{Defense Objectives} The goal of the defender is to protect against backdoor attacks in models. State-of-the-art defenses like CleanCLIP fine-tune poisoned models with extensive image-text pairs, which can be inefficient and impact accuracy. Our proposed strategy employs a targeted unlearning method, leveraging suspect datasets to selectively erase backdoor data, preserving model performance on clean data.

\subsection{Poisoned Sample Overfitting}
Faced with the challenge of ``weak'' backdoor shortcuts created by attackers, our defense strategy aims to further strengthen these shortcuts to better discover suspicious samples. To this end, we combine dataset analysis with a differentiated training approach, focusing on the segmentation of the poisoned dataset and strengthening the model's response to backdoor triggers through a specific training process.

We begin by dividing the dataset using a clean pre-trained model into suspicious $D_{\text{susb}}$ and clean $D_{\text{safe}}$ sample sets based on multimodal text similarity. In the reinforcement phase, we increase the suspicious set's cosine similarity, the model becomes more sensitive to backdoors, ensuring accurate trigger detection. Clean set $D_{\text{safe}}$ serves as a regularization for balance training, using InfoCE loss to prevent overfitting to clean samples, thus prioritizing the fitting of backdoor features. The process can be formulated as follows:
\begin{align*}
     \theta_{\text{overfitting}} = \min_{\theta} \Big\{ \frac{1}{\lvert D_{\text{susp}} \rvert}\sum_{i=1}^{\lvert D_{\text{susp}}\rvert}{\left[S(f_i(I^{\text{susp}}_i), f_t(T^{\text{susp}}_i))+1\right]^2} \\+ L_{\text{CLIP}}(D_{\text{safe}}) \Big\}
\end{align*}
where $f_i$ and $f_t$ denote the visual coder and text coder of the CLIP model, respectively, $S(I,T)$ denotes the cosine similarity of the image-text pair $(I,T)$, and $L_{\text{CLIP}}$ denotes the multimodal contrastive loss.

At this point, $D_{\text{susp}}$ and $D_{\text{safe}}$ represent the suspicious sample set and the clean dataset, respectively. With this staged and targeted training approach, we amplify the poisoning properties of the model, which helps pinpoint those samples that have the greatest impact on the model's security, comprising the oblivious subset used for backdoor defense.

\subsection{Suspicious Sample Detection}
We reanalyze the suspicious sample set using the overfitting poisoned model after enhancing the shortcuts and further perform a finer-grained backdoor analysis on the sample set. The goal of this process is to discover and localize the subsets of samples that have the greatest impact on backdoor oblivion, so that these backdoor features can be weakened or eliminated more effectively in subsequent processing, thereby improving the overall security and robustness of the model.

Specifically, we first compute, for each sample in the suspect sample set, its embedding features, which are generated by the poisoning model reinforcing the backdoor features, reflecting the multidimensional spatial location of the sample represented inside the poisoning model. Subsequently, we reorder the similarity scores of these embedded features and highly focus on the backdoor samples with the highest similarity scores. This can be represented as follows:
\begin{equation}
    D_{\text{topk}} = \left\{ (I_i, T_i) \in D_{\text{susp}} \mid \text{rank}(S(I_i, T_i)) \leq k \right\},
\end{equation}
where rank() denotes the similarity ranking of the image-text pair $(I,T)$ in the set, the higher the similarity, the smaller the rank value is.

Top-k ranked samples are more likely to carry backdoor triggers because they exhibit the highest activation scores compared to the other samples. This phenomenon suggests that when the model encounters these specific samples, the probability of the backdoor logic being activated is significantly higher, thus triggering a specific, predetermined response at the output layer of the model. By identifying these high similarity few-shot suspicious samples, we can not only focus on this small group of samples to effectively mitigate or eliminate the potential threat posed by backdoor attacks, but also reduce the overall cost of oblivious training.

\subsection{Token-level Local Unlearn}
To enhance our model's resilience against backdoor attacks, we introduce a targeted forgetting strategy that mitigates the attacks' impact without compromising model accuracy. This strategy focuses on selective, not wholesale, forgetting and preserving model knowledge while addressing the minimal, yet crucial modifications introduced by backdoors. Given the complexity of identifying specific regions for forgetting, especially with sophisticated attacks that seamlessly blend triggers, we opt for discrete text token forgetting. This approach, informed by the observation that backdoors less frequently distort text semantics, involves evaluating each token's contribution to backdoor effects, as outlined by ~\cite{Chefer_2021_ICCV_Mask}, and selectively forgetting less impactful ones. 

To further boost this process's efficiency, we employ data augmentation through Cartesian product combinations, enriching training data diversity. This method, known as token-based local forgetting, strategically strengthens the model against backdoor vulnerabilities.
\begin{align}
    \theta_{unlearn} &= \min_{\theta}({\frac{1}{\lvert D_{\text{unlearn}} \rvert}\sum^{\lvert D_{\text{unlearn}} \rvert}_{i=1}{S(I_i,T_i)}})
\end{align}
where $D_{\text{unlearn}}$, extended from $D_{\text{topk}}$ based on key insights, enhances the model's ability to forget backdoor samples efficiently, maintaining recognition of normal samples with minimized backdoor impact.

\section{Experiments}
\label{sec:experiments}
\textbf{Experimental Setting} A 500K subset of the CC3M dataset~\cite{sharma2018cc3m} and the CLIP model are used for backdoor attack experiments using ViT/32-B and Transformer as visual and text encoders. The experiment adds 1500 backdoor samples to this subset and employs four backdoor attack methods: BadNet, Blended, SIG, and SSBA. The model is poisoned and trained with a batch size of 128 and a learning rate of 1e-6 for 5 iterations. For backdoor defense, UBT first trains an overfitting poisoning model with a batch size of 64 and a learning rate of 1e-6 for 5 rounds of training, making it difficult to generalize to clean data. Then, UBT uses a forgetting technique to adjust the batch size to 64, the learning rate to 1e-5, and performs 3 rounds of training to eliminate backdoor feature memories from the model, enhancing security and robustness. The advanced CleanCLIP defense is used as a comparison method, and the specific experimental setup is described in ~\cite{bansal2023cleanclip}.

\begin{table}[t]
\caption{Backdoor defense results against different attacks.}
\label{tab:attack_results}
\scalebox{0.8}{%
\begin{tabular}{llll}
\toprule
Attack Method &  & CA & ASR \\
\midrule
Pretrained CLIP&  & 62.69 & - \\
\midrule
\multirow{4}{*}{BadNet} & No defense & 62.61 & 80.92\\ 
& CleanCLIP & 58.95 & 14.6\\ 
& UBT w/o token-level & 61.29 & 0.01 \\
& UBT  & 61.51 & 0.00 \\
\midrule
\multirow{4}{*}{Blended} & No defense & 62.58 & 97.99\\ 
& CleanCLIP & 59.43 & 2.24\\ 
& UBT w/o token-level & 60.81 & 0.156 \\
& GA + TEXT MASK (ours) & 60.56 & 0.08 \\
\midrule
\multirow{4}{*}{SIG} & No defense & 62.77 & 90.90\\ 
& CleanCLIP & 59.44 & 48.48\\ 
& UBT w/o token-level & 62.72 & 0.25 \\
& UBT & 62.70 & 0.27 \\
\midrule
\multirow{4}{*}{SSBA} & No defense & 62.77 & 66.22\\ 
& CleanCLIP & 58.90 & 15.53\\ 
& UBT w/o token-level & 62.20 & 4.332 \\
& UBT & 62.144 & 2.814 \\
\bottomrule
\end{tabular}%
}
\end{table}

\noindent\textbf{Backdoor Defense Results} Analyzing Tab.~\ref{tab:attack_results}, we draw the following conclusions:
1) The UBT defense strategy, especially the version with token-level technology, shows significant defense efficacy in all kinds of backdoor attack scenarios and is capable of effectively reducing the Attack Success Rate (ASR) to close to or completely zero, demonstrating its strong ability to defend against backdoor attacks. 2) When comparing the effectiveness of no defense, CleanCLIP defense, and different configurations of UBT (including the version without and with token level), it is clearly seen that the version of UBT using token level provides better protection in almost all cases, reduces the model's sensitivity to backdoor features, and enhances the model's security and robustness. 3) Additionally, even in scenarios with high attack success rates, such as combined attacks (97. 99\% ASR) and SIG attacks (90.90\% ASR), the UBT method still significantly reduces the effectiveness of attacks, proving its effectiveness as a backdoor defense.

\begin{figure}[t]
    \centering
    \includegraphics[width=0.45\textwidth]{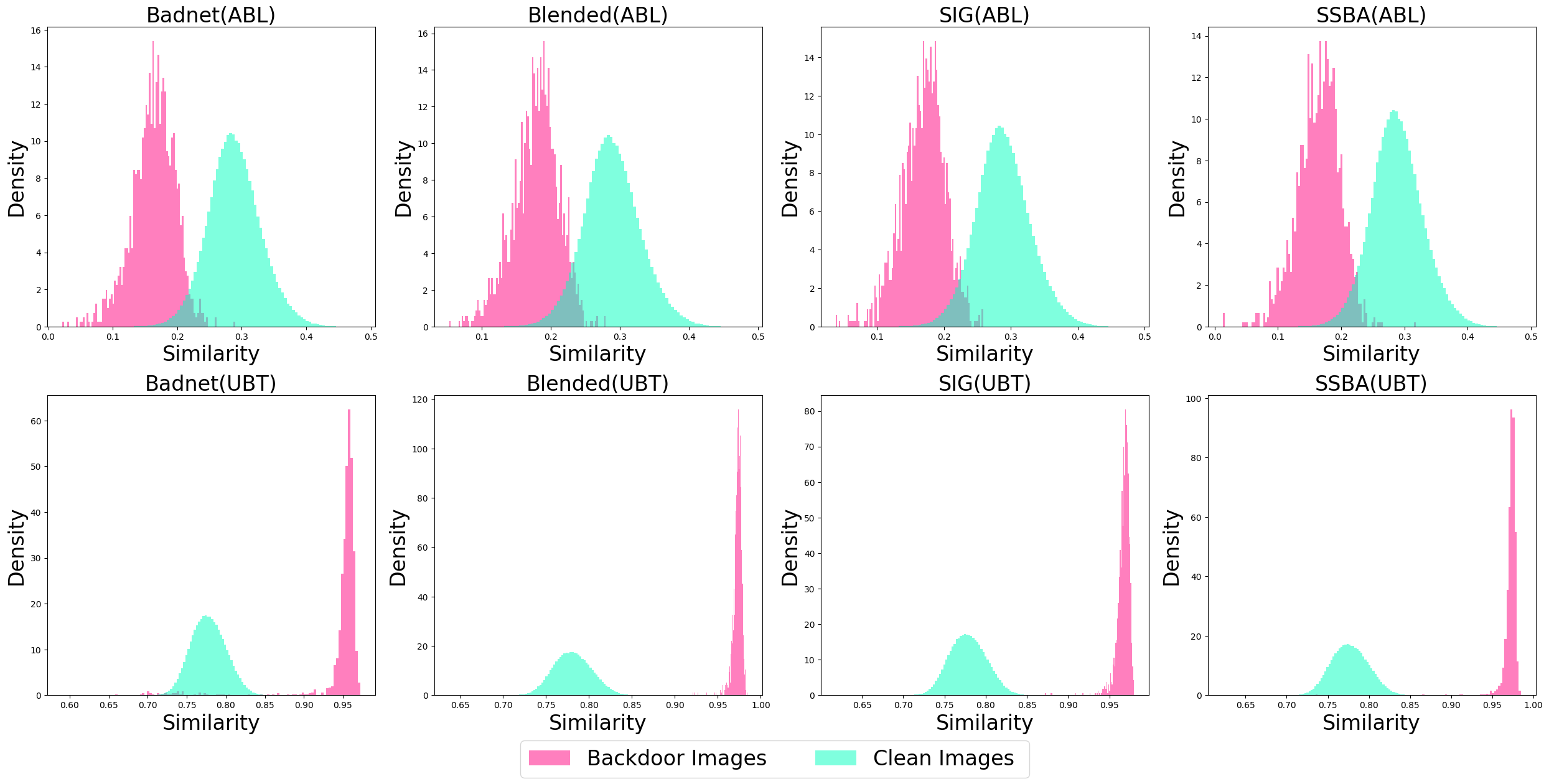}
    \caption{Sample distribution statistics under different defense methods.}
    \label{fig:ABL VS UBT}
\end{figure}

\noindent\textbf{Sample Separation Visualization} Based on Fig.~\ref{fig:ABL VS UBT}, we can draw two conclusions: 1) The UBT method significantly outperforms the ABL method in distinguishing backdoor images from clean images, as demonstrated by the clearly separated distributions at the bottom of the UBT graph. 2) This result indicates that UBT provides a more reliable defense mechanism because it can effectively reduce the overlap in similarity between clean images and backdoor images, thus improving the accuracy of security protection.
\section{Conclusion}
This study proposes a novel defense strategy against backdoor attacks in multimodal contrastive learning, achieving significant strides in nullifying backdoor shortcuts via few-shot poisoned pairs and token-level local unlearning. Our results not only confirm the method's efficacy in lowering attack success rates and ensuring model purification but also suggest avenues for future enhancements in optimization~\cite{li2022learning} and explanation~\cite{chen2024less}, further bolstering the security landscape of MCL.
{
    \small
    \bibliographystyle{ieeenat_fullname}
    \bibliography{main}
}


\end{document}